\documentclass[AMA,LATO1COL]{WileyNJD-v2}
\articletype{Article Type}%

\received{2020}
\revised{ 2020}
\accepted{ 2020}
\usepackage{moreverb}
\usepackage{mathtools}
\usepackage{amsfonts,latexsym}
\usepackage{caption}
\usepackage{subcaption}

\newcommand\BibTeX{{\rmfamily B\kern-.05em \textsc{i\kern-.025em b}\kern-.08em
T\kern-.1667em\lower.7ex\hbox{E}\kern-.125emX}}

\articletype{Article Type}%

\begin{document}

\title{Parallel framework for Dynamic Domain
Decomposition of Data Assimilation problems}










\abstract[Abstract]{We focus on PDE-based Data Assimilation problems (DA) solved by means of variational approaches and Kalman Filter algorithm. Recently, we presented a Domain Decomposition framework (we
call it DD-DA, for short) performing a decomposition of the whole physical domain  along
space and time directions, and joining the idea of Schwarz's methods and Parallel in Time (PinT)-
based approaches.
For effective parallelization of  domain decomposition algorithms,  the computational load assigned
to sub domains must be equally distributed. Usually  computational cost is proportional to the
amount of data entities assigned to partitions. Good quality partitioning also requires the volume
of communication during calculation to be kept at its minimum. In order to deal with DD-DA problems
where the observations are non uniformly distributed and general sparse, in the present work
we employ a parallel  load balancing algorithm - based on adaptive and dynamic defining of 
boundaries of  DD - which is aimed to balance workload according to data
location. We call it DyDD.  As the numerical model underlying  DA problems
arising from the so-called discretize-then-optimize approach is the  Constrained
Least Square model (CLS), we will use CLS as a reference state estimation problem and we validate
DyDD on different scenarios.}



\maketitle

\footnotetext{\textbf{Abbreviations:} DA, Data Assimilation; DD, Domain Decomposition; KF, Kalman Filter; CLS Constrained Least Square, DD-DA Domain Decomposition for Data Assimilation; DD-KF Domain Decomposition for Kalman Filter.}

\section{Introduction}\label{sec1}

Data Assimilation (DA, for short) encompasses the entire sequence of operations that, starting from observations/measurements of physical
quantities and from additional information - such as a mathematical model governing the evolution of these quantities - improve their estimate
minimizing inherent uncertainties. DA problems are usually formulated as an optimization problem where the objective function measures the
mismatch between the model predictions and the observed system states, weighted by the inverse of the error covariance matrices\cite{Cohn,Nichols}. In operational
DA the amount of observations is insufficient to fully describe the system and one cannot strongly rely on a data driven approach: the model
is paramount. It is the model that fills the spatial and temporal gaps in the observational network: it propagates information from observed to
unobserved areas. Thus, DA methods are designed to achieve the best possible use of a never sufficient (albeit constantly growing) amount of data,
and to attain an efficient data model fusion, in a short period of time. This poses a formidable computational challenge, and makes DA an example
of big data inverse problems\cite{JPP,WCEAS,PPAM2017,4DVAR}.
There is a lot of DA algorithms. Two main approaches gained acceptance as powerful methods: variational approach (namely 3DVAR and 4DVAR) and Kalman Filter (KF) \cite{Evensen,Kalman60,sorenson}. Variational approaches are based on the minimization of the objective function estimating the discrepancy between
numerical results and observations. These approaches assume that the two sources of information, forecast and observations, have errors that
are adequately described by stationary error covariances. In contrast to variational methods, KF (and its variants) is a recursive filter solving the
Euler-Lagrange equations. It uses a dynamic error covariance estimate evolving over a given time interval. The process is sequential, meaning that
observations are assimilated in chronological order, and KF alternates a forecast step, when the covariance is evolved, with an analysis step in
which covariance of the filtering conditional is updated. In both kind of methods the model is integrated forward in time and the result is used to
reinitialize the model before the integration continues. For any details interested readers are referred to\cite{Kalnay}.\\
\noindent Main operators of any DA algorithm are dynamic model and observation mapping. These are two main components of any variational approach and
state estimation problem, too. In this regard, in the following, as proof of concept of the DyDD framework, we start considering CLS model, seen as a
prototype of Data Assimilation model \cite{Gander}. CLS is obtained combining two overdetermined linear systems, representing the state and the observation
mapping, respectively. In this regards, in \cite{DD-KF} we presented a feaibility analysis on Constrained Least Square (CLS) models of an innovative Domain
Decomposition (DD) framework for using CLS in large scale applications. DD framework, based on Schwarz approach, that properly
combines localization and PDE-based model reduction inheriting the advantages of both techniques for effectively solving any kind of large scale
and/or real time KF application. It involves decomposition of the physical domain, partitioning of the solution, filter localization and model reduction,
both in space and in time.
There is a quite different rationale behind the DD framework and the so called Model Order Reduction methods (MOR) \cite{Rozier}, even though they are
closely related each other. The primary motivation of Schwarz based DD methods was the inherent parallelism arising from a flexible, adaptive and
independent decomposition of the given problem into several subproblems, though they can also reduce the complexity of sequential solvers. 
Schwarz Methods and theoretical frameworks are, to date, the most mature for this class of problems \cite{Ricerche2019,Gander,Schwarz}. MOR techniques are based on
projection of the full order model onto a lower dimensional space spanned by a reduced order basis. These methods has been used extensively in a
variety of fields for efficient simulations of highly intensive computational problems. But all numerical issues concerning the quality of approximation
still are of paramount importance \cite{Homescu}. As mentioned previously DD-DA framework makes it natural to switch from a full scale solver to a model
order reduction solver for solution of subproblems for which no relevant low-dimensional reduced space should be constructed. In the same way,
DD-DA framework allows to employ a model reduction in space and time which is coherent with the filter localization. In conclusion, main advantage
of the DD framework is to combine in the same theoretical framework model reduction, along the space and time directions, with filter localization,
while providing a flexible, adaptive, reliable and robust decomposition.
That said, any interest reader who wants to apply DD framework in a real-world application, i.e. with a (PDE-based) model state and an observation
mapping, once the dynamic (PDE-based) model state has been discretized, he should rewrite the state estimation problem under consideration
as a CLS model problem (cfr Section 3.1) and then to apply DD algorithm. In other words, she/he should follow the discretize-then-optmize
approach, common to most Data Assimilation problems and state estimation problems, before employing the DD framework.
Summarizing, main topics of DD framework can be listed as follows.

\begin{enumerate}
\item DD step: we begin by partitioning along space and time the domain into subdomains and then extending each subdomain to overlap its
neighbors by an amount. Partitioning can be adapted according to the availability of measurements and data.
\item  Filter Localization and MOR: on each subdomain we formulate a local DA problem analogous to the original one, combining filter localization
and model order reduction approaches.
\item  Regularization constraints: in order to enforce the matching of local solutions on the overlapping regions, local DA problems are slightly modified
by adding a correction term. Such a correction term balances localization errors and computational efforts, acting as a regularization
constraint on local solutions. This is a typical approach for solving ill posed inverse problems (see for instance 33).
\item Parallel in Time: as the dynamic model is coupled with DA operator, at each integration step we employ, as initial and boundary values of all
reduced models, estimates provided by the DA model itself, as soon as these are available.
\item Conditioning: localization excludes remote observations from each analyzed location, thereby improving the conditioning of the error
covariance matrices.
To the best of our knowledge, such ab-initio space and time decomposition of DA models has never been investigated before. A spatially distributed
KF into sensor based reduced-order models, implemented on a sensor networks where multiple sensors are mounted on a robot platform for target
tracking, is presented in \cite{Battistelli,Kahn}.
\end{enumerate}

\subsection{Contribution of the present work} 
In this work we focus on the introduction of a dynamic redefining of the DD, aimed  to  efficiently deal with DA problems where  observations are non
uniformly distributed and general sparse. Indeed,  in such cases a static and/or a-priori DD strategy could not ensure a well balanced workload,
while a way to re-partition the domain so that  subdomains maintain a nearly equal number of observations plays an essential role in the success
of any effective DD approach. We present a revision of DD framework such that a dynamic load balancing algorithm allows for a minimal data
movement restricted to the neighboring processors. This is achieved by considering a connected graph induced by the domain partitioning whose
vertices represent a subdomain associated with a scalar representing the number of observations on that sub domain. Once the domain has been
partitioned, a load balancing schedule (scheduling step) should make the load on each subdomain equals to the average load providing the amount of
load to be sent to adjacent subdomains (migrations step). The most intensive kernel is the scheduling step which defines a schedule for computing
the load imbalance (which we quantify in terms of number of observations) among neighbouring subdomains. Such quantity is then used to update
the shifting the adjacent boundaries of sub domains (Migration step) which are finally re mapped to achieve a balanced decomposition. We are assuming that load
balancing is restricted to the neighbouring domains so that we reduce the overhead processing time. Finally, following \cite{dynamicload} we  use a diffusion
type scheduling algorithm minimizing the euclidean norm of data movement. The resulting constrained optimization problem leads to normal equations whose matrix is associated to the decomposition graph. The disadvantage is that the overhead time, due to the final balance between sub domains, strongly  depends  on the degree of the vertices of processors graph, i.e. on the number of
neighbouring subdomains for each subdomain. Such overhead represents the surface-to-volume ratio whose impact on the overall performance of the parallel algorithm decreases as the problem size increases. 

\subsection{Related Works} 
There has been widespread interests in load balancing since the introduction of large scale multiprocessors. Applications requiring dynamic load
balancing mainly include  parallel solution of a partial differential equation (PDE) by finite elements on an unstructured grids \cite{Diniz} or  parallelized particle
simulations \cite{efficienza}.
Load balancing is one of the central problems which have to be solved in designing  parallel  algorithms. Moreover, problems whose workload changes during the
computation or it depends on data layout which may be unknown a priori, will necessitate the redistribution of the data in order to retain efficiency.
Such a strategy is known as dynamic load balancing. Algorithms for dynamic load balancing, as in \cite{Cybenco, Boillat,Xu1, Xu2}, are based on transferring an amount of work
among processors to neighbours; the process is iterated until the load difference between any two processors is smaller than a specified value, consequently
it will not provide a balanced solution immediately.  A multilevel
diffusion method for dynamic load balancing \cite{Horton}, is based on bisection of processor graph. The disadvantage is that, to ensure the connectivity of subgraphs, movement of data
between non-neighbouring processors can occur.    The mentioned algorithms do not take into
account one important factor, namely that the data movement resulting from the load balancing schedule should be kept to minimum.

\subsubsection{Organization of the work}
The present work is organized as follows. As we apply the proposed framework to CLS model which can be seen as prototype of variational DA models,  in order to improve the readability of the article, in Section \S 2 we give a brief overview of DA methods, i.e. both Kalman Filter and Variational DA, the variational formulation of KF and finally  we give a brief description of  CLS model. 
In Section \S 3 we describe main features of DD framework and its application to CLS model. DyDD is presented in Section \S 4, through a
graphical description and the numerical algorithm. Validation and performance results are presented in Section \S 5 and, finally, in Section \S 6 we give conclusions and future works.

\section{The Background}
In order to improve the readability of the article, in this section we give a brief overview of DA methods, i.e. both Kalman Filter and Variational DA, then we review CLS model as prototype of DA models. To this end, we also review the variational formulation of KF, i.e. the so-called VAR--KF formulation, obtained minimizing the sum of the weighted Euclidean norm of the model error  and the weighted Euclidean norm of the observation error.

\subsection{Kalman Filter (KF)}

\noindent Given $x_{0}\in \mathbb{R}^{n}$, let $x(t)\in \mathbb{R}^{n}$, $\forall t\in [0,T]$,  denote  the state of a dynamic system governed by the mathematical model $\mathcal{M}_{t,t+\Delta t}[ x(t)]$, $\Delta t>0$:
\begin{equation}\label{modello}
\left\{\begin{array}{ll}
x(t+ \Delta t)&=\mathcal{M}_{t, t+ \Delta t}(x(t)), \ \ \forall t,t+\Delta t \in [0,T]\\
x(0)&=x_{0}
\end{array}, \right.
\end{equation}  
and let:
\begin{equation}\label{osservazioni}
y(t+\Delta t)=\mathcal{H}_{t+\Delta t}[x(t+\Delta t)],
\end{equation}
denote the observations where  $\mathcal{H}_{t+\Delta t}$ is the observations mapping.
Chosen $r\in \mathbb{N}$, we consider $r+2$ points in $[0,T]$ and $\Delta t=\frac{T}{r+1}$.\\ Let $\{t_{k}\}_{k=0,1,\ldots,r+1}$ be a discretization of $[0,T]$, where $t_{k}=k\Delta t$, and let $\widehat{x}_{k}$ be  the state estimate at time $t_{k}$, for $k=1,...,r+1$; we will use  the following operators  \cite{sorenson}: , for $k=0,1,...,r$,  $M_{k,k+1}\in \mathbb{R}^{n\times n}$, denoting the discretization of a linear approximation of $\mathcal{M}_{t_{k},t_{k+1}}$ and for $k=0,1,...,r+1$,
 $H_{k}\in \mathbb{R}^{m\times n}$ which is the discretization of a linear approximation of $\mathcal{H}_{t}$ with $m>n$. Moreover, we let 
 $w_{k}\in \mathbb{R}^{n}$ and $v_{k}\in \mathbb{R}^{m}$ be  model and observation errors with normal distribution and zero mean such that $E[w_{k}v_{i}^{T}]=0$, for $i,k=0,1,...,r+1$, where $E[\cdot]$ denotes the expected value;
$Q_{k}\in \mathbb{R}^{n\times n}$ and $R_{k}\in \mathbb{R}^{m \times m}$, are   covariance matrices of the errors on the model and on the observations, respectively i.e.
\begin{equation*}
Q_{k}:=E[w_{k}w_{k}^{T}] \quad R_{k}:=E[v_{k}v_{k}^{T}] \quad \textit{$\forall$ $k=0,1,...,r+1$}.
\end{equation*}
These matrices are symmetric and positive definite.

\noindent \textbf{KF method}: 
KF method consists in calculating the estimate $\widehat{x}_{k+1}$, at time $t_{k+1}$, of the state $x_{k+1}\in \mathbb{R}^{n}$:
\begin{equation}\label{sistema_kalmandiscreto}
x_{k+1}=M_{k,k+1}x_{k}+w_{k}, \quad \textit{$\forall k=0,1,...,r$}
\end{equation}
such that 
\begin{equation}\label{problema1}
y_{k+1}=H_{k+1}{x}_{k+1}+v_{k+1},\quad \textit{$\forall k=0,1,...,r$}.
\end{equation}

\noindent \textbf{KF algorithm}: \label{def_kalman} Given $\widehat{x}_{0}\in \mathbb{R}^{n}$ and $P_{0}=O\in \mathbb{R}^{n\times n}$ a null matrix, for each $k=0,1,\ldots,r$  KF algorithm is made  by two main operations: the  Predicted phase, consisting of
the computation of the  predicted state estimate:
\begin{equation}\label{stimapredetta}
x_{k+1}=M_{k,k+1}\widehat{x}_{k};
 \end{equation}
and of the  predicted error covariance matrix: 
\begin{equation}\label{predictedmatrix}
P_{k+1}=M_{k,k+1}P_{k}M_{k,k+1}^{T}+Q_{k} ;
\end{equation}
and the Corrector phase, consisting of the computation of Kalman gain:
\begin{equation}\label{guadagnodikalman}
K_{k+1}=P_{k+1}H_{k+1}^{T}(H_{k+1}P_{k+1}H_{k+1}^{T}+R_{k+1})^{-1},
\end{equation}
of  Kalman covariance matrix:
\begin{equation*}
P_{k+1}=(I-K_{k+1}H_{k+1})P_{k+1},
\end{equation*}
and  of   Kalman state estimate:
\begin{equation}\label{stimastato}
\widehat{x}_{k+1}=x_{k+1}+K_{k+1}(y_{k+1}-H_{k+1}x_{k+1}).
\end{equation}

\noindent Finally, we introduce the VAR-KF model. 
For  $k=0,1,\ldots,r$:
\begin{equation*}
\begin{array}{ll}
\widehat{x}_{k+1}&=argmin_{x_{k+1}\in \mathbb{R}^{n}}J_{k+1}(x_{k+1})\\&=argmin_{x_{k+1}\in \mathbb{R}^{n}}\left\{||x_{k+1}-M_{k,k+1}\widehat{x}_{k}||_{Q_{k}}^{2}+||y_{k+1}-H_{k+1}{x}_{k+1}||_{R_{k+1}}^{2}\right\}.
\end{array}
\end{equation*}

\section{VAR DA model Set Up }
If $\Omega \subset \mathbb{R}^{n}$, $n\in \mathbb{N}$,  is a spatial  domain with a Lipschitz boundary, let:
\begin{equation}\label{modelloDA}
\left\{ \begin{array}{ll}
u(t+h,x)=\mathcal{M}_{t,t+h}[u(t,x)] & \textrm{$\forall x \in \Omega$, $t,t+h \in [0,T]$, $(h >0)$} \\
u(t_{0},x)=u_{0}(x) & \textrm{$ t_{0}\equiv 0, \ \ x\in \Omega$}\\
u(t,x)=f(x) & \textrm{$ x\in \partial \Omega$, $\forall t \in [0,T]$}\\
\end{array}, \right.
\end{equation}
be a symbolic description of the 4D--DA model of interest where
$$u:(t,x) \in [0,T] \times \Omega \mapsto u(t,x)= [u[1](t, x), u[2](t, x),\ldots, u[pv](t, x)],
$$
is the state function of $\mathcal{M}$ with $pv\in \mathbb{N}$ the number of physical variables, $f$ is a known function defined on the boundary $ \partial \Omega$, and let
$$v:(t,x) \in [0,T] \times \Omega \mapsto v(t,x),$$
be the observations function, and
$$\mathcal{H}: u(t,x) \mapsto v(t,x), \ \ \ \ \ \forall (t,x) \in [0,T] \times \Omega,$$
denote the non-linear observations mapping. To simplify future treatments we assume $pv \equiv 1$. We consider $N_{p}$ points of $\Omega \subset \mathbb{R}^{n}$ $:$ $\{x_{j}\}_{j=1,\ldots,N_{p}}\subset \Omega$; $n_{obs}$ points of $\Omega$, where $n_{obs} <<N_{p}$, $:$ $\{y_{j}\}_{j=1,\ldots,n_{obs}}$;
 $N$ points of [0,T] $:$ $D([0,T])=\{t_{l}\}_{l=0,1,\ldots,N-1}$ with $t_{l}=t_{0}+l(h T)$; the vector 
$$u_{0}=\{u_{0,j}\}_{j=1,\ldots,N_{p}}\equiv \{u_{0}(x_{j})\}_{j=1,\ldots,N_{p}} \in \mathbb{R}^{N_{p}},$$
which is the state at time $t_{0}$; the operator 
$$M_{l-1,l}\in \mathbb{R}^{N_{p} \times N_{p}}, \ \ \ l=1,\ldots,N,$$
representing a discretization of a linear approximation of $\mathcal{M}_{t_{l-1},t_{l}}$ from $t_{l-1}$ to $t_{l}$;   the vector $b\in \mathbb{R}^{N_{p}}$ accounting boundary conditions; the vector 
$$
u^{b}:=\{u_{l,j}^{b}\}_{l=1,\ldots,N-1;j=1,\ldots,N_{p}} \equiv \{u^{b}(t_{l},x_{j})\}_{l=1,\ldots,N-1;j=1,\ldots,N_{p}} \in \mathbb{R}^{N_{p}\cdot (N-1)},$$
representing solution of $M_{l-1,l}$ at $t_{l}$ for $l=1,\ldots,N$, i.e. the background; the vector
$$v_{l}\equiv \{v(t_{l},y_{j})\}_{j=1,\ldots,n_{obs}}\in \mathbb{R}^{l\cdot n_{obs}},$$
consisting of  observations at $t_{l}$, for $l=0,\ldots,N-1$; the linear operator
$$H_{l}\in \mathbb{R}^{n_{obs} \times N_{p}}, \ \ \ l=0,\ldots,N-1,$$
representing a linear approximation of $\mathcal{H}$; matrix $G\equiv G_{N-1}\in \mathbb{R}^{(N \cdot n_{obs})\times N_{p} }$ such that
\begin{displaymath}\label{mat_G}
G_{l}=\left\{ \begin{array}{ll}
\left[\begin{array}{ll}
H_{0}\\ H_{1}\\ \vdots \\H_{l-1} \end{array}\right] & \textrm{$l>1$} \\
\\
H_{0}& \textrm{$ l=1$}
\end{array}, \right.
\end{displaymath} and $\textbf{R}=diag(\textbf{R}_{0},\textbf{R}_{1},\ldots,\textbf{R}_{N-1})$  and \textbf{Q}$=\textbf{V}\textbf{V}^{T}$, covariance matrices of the errors on  observations and  background, respectively.
We now define the 4D--DA inverse problem \cite{Fourier2002}.
\begin{definition}(The 4D-DA inverse problem). Given the vectors
$v=\{v_{l}\}_{l=0,\ldots,N-1} \in \mathbb{R}^{N \cdot n_{obs}}, \ \ u_{0} \in \mathbb{R}^{N_{p}},$
and the block matrix
$G \in \mathbb{R}^{(N \cdot n_{obs}) \times N_{p} },$
a 4D--DA problem concerns the computation of
$$u^{DA} \in \mathbb{R}^{N_{p}},$$
such that
\begin{equation}\label{DA}
v=G\cdot u^{DA},
\end{equation}
subject to the constraint that
$u_{0}^{DA}=u_{0}.$
\end{definition}  
\noindent We also introduce the regularized 4D-DA inverse problem, i.e. the  4D--VAR DA problem.
\begin{definition}(The 4D--VAR DA  problem). The 4D--VAR DA  problem concerns the computation of:
\begin{equation}\label{varDA}
u^{DA}=argmin_{u\in \mathbb{R}^{N_{p}}}J(u),
\end{equation}
with
\begin{equation}\label{funzionale}
J(u)=\alpha ||u-u^{b}||_{\textbf{Q}^{-1}}^{2}+||Gu-v||_{\textbf{R}^{-1}}^{2},
\end{equation}
where $\alpha$ is the regularization parameter.
\end{definition}

\noindent \emph{Remark}: It is worth noting that here we are considering a linear approximation of the observation operator, hence a linear operator $G$, although this is not at all required, at least in the formulation of the 4D--VAR problem. A more general  approach for numerically linearize and solve  4D--VAR DA  problem  consists in defining a sequence of local approximations of $\mathbf{J}$ where each member of the sequence is minimized by  employing Newton's  method or one its variants. More precisely, two approaches could be employed:
\begin{enumerate}
  \item[(a)] by truncating   Taylor's series expansion of $\mathbf{J}$ at the second order, 
    giving a quadratic approximation of $\mathbf{J}$, let us say $\mathbf{J}^{QN}$. Newton'methods (including LBFGS and Levenberg-Marquardt) use $\mathbf{J}^{QD}$. The minimum is computed solving the linear system involving the Hessian matrix $\nabla ^2\mathbf{{J}}$, and the negative gradient $-\nabla \mathbf{{J}}$.
  \item[(b)]  by truncating Taylor's series expansion of $\mathbf{J}$ at the first order  which gives a linear approximation of $\mathbf{J}$, let us say let us say $\mathbf{J}^{TL}$. Gauss-Newton's methods (including Truncated or Approximated Gauss-Newton uses  $\mathbf{J}^{TL}$). The minimum is computed  solving the normal equations arising from  the local Linear Least Squares  problem. 
\end{enumerate}
Both approaches will employ the Tangent Linear Model (TLM) and the adjoint operator of the observation mapping and of the model of interest \cite{Cacuci}. \\ 

\noindent \emph{Remark}: 
Computational kernel of  variational approaches (namely 3D-Var and 4D-Var)  is a linear system, generally solved by means of  iterative methods; the  iteration matrix is related to matrix \textbf{Q}, which usually has a Gaussian correlation structure \cite{4DVAR}. Matrix \textbf{Q} can be written in the form \textbf{Q}$=\textbf{V}\textbf{V}^{T}$, where \textbf{V} is the square root of  \textbf{Q}, namely it is a Gaussian matrix.
As a consequence,  products of \textbf{V} and a vector \textbf{z} are Gaussian convolutions which can be efficiently computed by applying Gaussian recursive filters as in \cite{GaussianF}.  \\
\noindent In our case study we carry out on CLS model,  we apply KF and DD-KF to  CLS model, then in this case it results that matrix $\textbf{Q}$ is the null matrix while  matrix $\textbf{R}$ is diagonal \cite{DD-KF}.

\subsection{Constrained Least Squares (CLS) Problem}

\noindent Let 
\begin{equation}\label{sistema0}
H_{0}x_{0}=y_{0}, \quad  H_{0}\in \mathbb{R}^{m_{0}\times n}, \quad  y_{0}\in \mathbb{R}^{m_{0}}, \quad x_{0}\in \mathbb{R}^{n} 
\end{equation}
be the overdetermined linear system (the state), where $rank(H_{0})=n>0$, $m_{0}>n$. \\
\noindent Given  $H_{1}\in \mathbb{R}^{ m_{1} \times n}$, $y_{1}\in \mathbb{R}^{m_{1}}$, $x_{1}\in \mathbb{R}^{n}$, $x\in \mathbb{R}^{n}$ (the observations), we consider the system 
\begin{equation}\label{sistema_kalman}
S: \ \ Ax=b
\end{equation}
where
\begin{equation}\label{matrice}
A=\left[ \begin{array}{llll}
H_{0}  \\
H_{1} \\
\end{array} \right]\in \mathbb{R}^{(m_{0}+ m_{1}) \times n}, \quad 
 b=\left[ \begin{array}{llll}
y_{0}   \\
y_{1}\\
\end{array} \right] \in \mathbb{R}^{m_{0}+ m_{1}},
\end{equation}
and $m_{1}>0$.
Let $R_{0}\in \mathbb{R}^{m_{0}\times m_{0}}$, $R_{1}\in \mathbb{R}^{m_{1}\times m_{1}}$ be  weight matrices and $R=diag(R_{0},R_{1})\in \mathbb{R}^{(m_{0}+m_{1}) \times (m_{0}+m_{1})}$. \\
\noindent CLS problem consists in the computation of $\widehat{x}$ such that:
\begin{equation}\label{minimi_quadrati}
CLS: \  \ \widehat{x}=argmin_{x\in \mathbb{R}^{n}}J(x)
\end{equation}
with
\begin{equation}\label{operatore}
\begin{array}{ll}
J(x)=||Ax-b||_{R}^{2}
=||H_{0}x-y_{0}||_{{R}_{0}}^{2}+||H_{1}x-y_{1}||_{{R}_{1}}^{2},
\end{array} 
\end{equation}
where $\widehat{x}$ is given by
\begin{equation}\label{eq_normali}
(A^{T}RA)\widehat{x}=A^{T}Rb \Rightarrow \ \widehat{x}=(A^{T}RA)^{-1}A^{T}Rb
\end{equation}
or, 
\begin{equation}\label{sol_normal}
\widehat{x}=(H_{0}^{T}R_{0}H_{0}+H_{1}^{T}R_{1}H_{1})^{-1}(H_{0}^{T}R_{0}y_{0}+H_{1}^{T}R_{1}y_{1}).
\end{equation}
We refer to $\widehat{x}$ as  solution in least squares sense of system in (\ref{sistema_kalman}).

\noindent \emph{Remark}:
  Besides covariance matrices of  errors, main components of KF algorithm are dynamic model and observation mapping. These are two
main components of any variational Data Assimilation operator and state estimation problem, too. In this regard, in the following, as proof of concept
of  DD framework, we start considering CLS model as a prototype of a variational Data Assimilation model, at a given time. CLS is obtained
combining two overdetermined linear systems, representing the state and the observation mapping, respectively. Then, we introduce  VAR-KF
method as reference data sampling method solving CLS model problem. VAR-KF will be decomposed by using the proposed DD framework. That
said, any interest reader who wants to apply DD framework in a real-world application, i.e. with a (PDE-based) model state and an observation
mapping, once the dynamic (PDE-based) model state has been discretized, he should rewrite the state estimation problem under consideration as
a CLS model problem (cfr Section 3.1) and then to apply CLS algorithm. In other words, she/he should follow the discretize-then-optmize approach,
common to most Data Assimilation problems and state estimation problems, before employing  DyDD framework.

\section{DD-framework}
As DyDD is the refinement of initial DD, in   the following we first  give  mathematical settings useful to define the domain 
decomposition framework. Then, in next section we focus on DyDD. 
\subsection{DD set up}
\begin{definition} \textbf{(Matrix Reduction)}
Let $B=[B^{1} \ B^{2} \ \dots \ B^{n}] \in \mathbb{R}^{m \times n}$ be a matrix with $m,n\ge 1$ and $B^{j}$ the $j-th$ column of $B$ and $I_j=\{1,\dots,j\}$ and $I_{i,j}=\{i,\dots,j\}$ for $i=1,\dots,n-1$;  $j=2,\dots,n$, and $i<j$ for every $(i,j)$.  Reduction  of $B$ to $I_j$ is:
\begin{equation*}
|_{I_{j}}: \ B\in \mathbb{R}^{m\times n} \rightarrow \ B|_{ I_{j}}=[B^{1} \ B^{2} \ \dots \ B^{j}]\in \mathbb{R}^{m \times j}, \quad \textit{ $j=2,\dots,n$},
\end{equation*}
and to $I_{i,j}$
\begin{equation*}
|_{ I_{i,j}}: \ B\in \mathbb{R}^{m\times n} \rightarrow \ B|_{ I_{i,j}}=[B^{i} \ B^{i+1} \ \dots \ B^{j}]\in \mathbb{R}^{m \times j-i}, \quad \textit{ $i=1,\dots,n-1$, $j>i$},
\end{equation*}
where $B|_{ I_{j}}$ and $B|_{ I_{i,j}}$ denote reduction of  $B$ to $I_{j}$ and $I_{i,j}$, respectively.
\end{definition}
\begin{definition}  \textbf{(Vector Reduction)}
Let $w=[w_{t} \ w_{t+1} \ \dots \ w_{n}]^{T} \in \mathbb{R}^{s}$ be a vector with $t\ge 1$, $n>0$, $s=n-t$ and $I_{1,r}=\{1,\dots,r\}$, $r>n$ and  $n>t$. The extension  of $w$  to  $I_{r}$ is:
\begin{equation*}
EO_{I_{r}}: \ w\in \mathbb{R}^{s} \rightarrow \ EO_{I_{r}}(w)=[\bar{w}_{1} \ \bar{w}_{2} \ \dots \ \bar{w}_{r}]^{T} \in \mathbb{R}^{r},
\end{equation*}
where  $i=1,\ldots,r$ 
\begin{equation*}
\bar{w}_{i} =\left\{ \begin{array}{ll} 
w_{i} & \textit{if $t\le i\le n$}\\
0 & \textit{if $i> n$ and $i<t$}\\
\end{array} \right. \quad .
\end{equation*}
\end{definition}
\noindent We  introduce  reduction of  $J$, as given  in (\ref{operatore}).
\begin{definition} \textbf{(Model Reduction)}
Let us consider  $A\in \mathbb{R}^{(m_{0}+m_{1})\times n}$, $b\in \mathbb{R}^{m_{0}+m_{1}}$,  the matrix and the vector defined in (\ref{matrice}),  $I_{1}=\{1,\ldots,n_{1}\}$, $I_{2}=\{1,\ldots,n_{2}\}$ with $n_{1},n_{2}>0$ and the vectors $x\in \mathbb{R}^{n}$. Let
\begin{equation*}
J|_{(I_{i},I_{j})}:(x|_{I_{i}},x|_{I_{j}})\longmapsto J|_{(I_{i},I_{j})}(x|_{I_{i}},x|_{I_{j}})\quad \textit{$\forall i,j=1,2$}
\end{equation*}
denote the reduction  of $J$ defined in (\ref{operatore}). It is defined as
\begin{equation}\label{operatore_ris}
J|_{(I_{i},I_{j})}(x|_{I_{i}},x|_{I_{j}})=||H_{0}|_{I_{i}}x|_{I_{i}}-(y_{0}+H_{0}|_{I_{j}}x|_{I_{j}})||_{R_{0}}^{2}+||H_{1}|_{ I_{i}}x|_{I_{i}}-(y_{1}+H_{1}|_{ I_{j}}x|_{I_{j}})||_{R_{1}}^{2},
\end{equation}
for $i,j=1,2$.
\end{definition}
\noindent For simplicity of notations we let $J_{i,j}\equiv J|_{(I_{i},I_{j})}$.

\subsection{DD-CLS problems: DD  of CLS model}

\noindent We apply  DD approach for solving  system  $S$ in (\ref{sistema_kalman}). Here, for simplicity of notations, we consider two subdomains.

\begin{definition} (DD-CLS model\cite{DD-KF}) Let $S$ be the overdetermined linear system in (\ref{sistema_kalman}) and $A\in \mathbb{R}^{(m_{0}+m_{1})\times n}$, $b\in \mathbb{R}^{m_{0}+m_{1}}$ the matrix and the vector defined in (\ref{matrice}) and ${R}_{0}\in \mathbb{R}^{m_{0}\times m_{0}}$, ${R}_{1}\in \mathbb{R}^{m_{1}\times m_{1}}$, $R=diag(R_{0}, {R}_{1})\in \mathbb{R}^{(m_{0}+m_{1})\times (m_{0}+m_{1})}$ be the weight matrices with $m_{0}>n$ and $m_{1}>0$.
Let us consider  the index set of columns of $A$, $I=\{1,\dots,n\}$. DD-CLS model consists of: 
\begin{itemize}
\item DD step. It consists of DD of $I$: 
\begin{equation}\label{I1I2}
I_{1}=\{1,\ldots,n_{1}\}, \ \ \ I_{2}=\{n_{1}-s+1,\ldots,n\},
\end{equation}
where $s\ge 0$ is the number of indexes in common, $|I_{1}|=n_{1}>0$, $|I_{2}|=n_{2}>0$, and the overlap sets
\begin{equation}\label{set_disgunti1}
{I}_{1,2}=\{{n}_{1}-s+1,\ldots,n_{1}\},
\end{equation}
If $s=0$, then $I$ is decomposed  without using the overlap, i.e. $I_{1}\cap I_{2}=\emptyset$ and ${I}_{1,2}\neq \emptyset$, instead if $s>0$ i.e.  $I$ is decomposed using  overlap, i.e. $I_{1}\cap I_{2}\neq \emptyset$ and ${I}_{1,2}=\emptyset$; 
restrictions of  $A$ to  $I_{1}$ and $I_{2}$ defined in (\ref{I1I2})
\begin{equation}\label{restrizione_mat}
{A}_{1}=A|_{ {I}_{1}}\in \mathbb{R}^{(m_{0}+m_{1})\times n_{1}},\quad {A}_{2}=A|_{  {I}_{2}}\in \mathbb{R}^{(m_{0}+m_{1})\times n_{2}},
\end{equation} 

\item  DD-CLS step:
given $x_{2}^{0}\in \mathbb{R}^{n_{2}}$,  according to the ASM (Alternating Schwarz Method) in \cite{Gander},   DD-CLS approach consists in solving for $n=0,1,2,\ldots$ the following overdetermined linear systems:
\begin{equation}\label{2sistemi}
S_{1}^{n+1}: \ \ {A}_{1}x_{1}^{n+1}={b}-A_{2}x_{2}^{n}; \quad S_{2}^{n+1}: \ \  {A}_{2}x_{2}^{n+1}={b}-A_{1}x_{1}^{n+1},
\end{equation}
 by employing a regularized VAR-KF model. It means that DD-CLS consists of  a sequence of two subproblems:
\begin{equation}\label{P1}
\begin{array}{ll}
P_{1}^{n+1}: \ \widehat{x}_{1}^{n+1}&=argmin_{x_{1}^{n+1}\in \mathbb{R}^{n_{1}}}J_{1}(x_{1}^{n+1},x_{2}^{n})\\&=argmin_{x_{1}^{n+1}\in \mathbb{R}^{n_{1}}}\left[J|_{(I_{1},I_2)}(x_{1}^{n+1},x_{2}^{n})+\mu \cdot \mathcal{O}_{1,2}(x_{1}^{n+1},x_{2}^{n})\right]
\end{array}
\end{equation}
\begin{equation}\label{P2}
\begin{array}{ll}
P_{2}^{n+1}: \ \widehat{x}_{2}^{n+1}&=argmin_{x_{2}^{n+1}\in \mathbb{R}^{n_{2}}}J_{2}(x_{2}^{n+1},x_{1}^{n+1})\\&=argmin_{x_{2}^{n+1}\in \mathbb{R}^{n_{2}}}\left[J|_{(I_{2},I_1)}(x_{2}^{n+1},x_{1}^{n+1})+\mu \cdot \mathcal{O}_{1,2}(x_{2}^{n+1},x_{1}^{n+1})\right]
\end{array}
\end{equation}
where ${I_{i}}$ is defined in (\ref{I1I2}) and $J|_{I_{i},I_{j}}$ is defined  in (\ref{operatore_ris}), $\mathcal{O}_{1,2}$  is the overlapping operator and $\mu>0$  is the regularization parameter.
\end{itemize}
\end{definition}
\begin{remark}
If   $I$  is decomposed without using  overlap (i.e. $s=0$), then $\widehat{x}_{1}^{n+1}\in \mathbb{R}^{n_{1}}$ and $\widehat{x}_{2}^{n+1}\in \mathbb{R}^{n_{2}}$ can be written in terms of normal equations as follows
\begin{equation}\label{sistemi_norm}
\begin{array}{ll}
\tilde{S}_{1}^{n+1}: \ (A_{1}^{T}RA_{1})\widehat{x}_{1}^{n+1}=A_{1}^{T}R(b-A_{2}x_{2}^{n}) &\Rightarrow \widehat{x}_{1}^{n+1}=(A_{1}^{T}RA_{1})^{-1}A_{1}^{T}Rb_{1}^{n} \\
\tilde{S}_{2}^{n+1}: \ (A_{2}^{T}RA_{2})\widehat{x}_{2}^{n+1}=A_{2}^{T}R(b-A_{1}x_{1}^{n+1}) &\Rightarrow \widehat{x}_{2}^{n+1}=(A_{2}^{T}RA_{2})^{-1}A_{2}^{T}Rb_{2}^{n+1},
\end{array}
\end{equation}
where $b_{1}^{n}=b-A_{2}x_{2}^{n}$ and $b_{2}^{n+1}=b-A_{1}x_{1}^{n+1}$.
\end{remark}
\begin{remark} Regarding the  operator $\mathcal{O}_{1,2}$, we consider  $x_{1}\in \mathbb{R}^{n_{1}}$ and $x_{2}\in \mathbb{R}^{n_{2}}$,   and we pose
\begin{equation*}
\mathcal{O}_{1,2}(x_{i},x_{j})=||EO_{I_{i}}({x}_{i}|_{{I}_{1,2}})-EO_{I_{i}}({{x}_{j}}|_{{I}_{1,2}})||, \ \textit{$i,j=1,2$}
\end{equation*} 
with $EO_{I_{i}}({x}_{1}|_{I_{1,2}})$, $EO_{I_{i}}({x}_{2}|_{I_{1,2}})$ be  the extension  to  $I_{i}$, of restriction to  ${I}_{1,2}$  in (\ref{set_disgunti1}) of $x_{1}\in \mathbb{R}^{n_{1}}$ and $x_{2}\in \mathbb{R}^{n_{2}}$, respectively. Operator $\mathcal{O}_{1,2}$ represents the exchange of data on the overlap  $I_{1,2}$ in (\ref{set_disgunti1}). 
\end{remark}

\normalsize
\begin{remark} DD-CLS gives to sequences  $\{x^{n+1}\}_{n\in \mathbb{N}_{0}}$:
\begin{equation}\label{successione}
x^{n+1}=\left\{\begin{array}{ll}
\widehat{x}_{1}^{n+1}|_{I_{1}\setminus I_{1,2}}& \textit{on $I_{1}\setminus I_{1,2}$}\\
\frac{\mu}{2}(\widehat{x}_{2}^{n+1}|_{I_{1,2}}+\widehat{x}_{1}^{n+1}|_{I_{1,2}})& \textit{on $I_{1,2}$}\\
\widehat{x}_{2}^{n+1}|_{I_{2}\setminus I_{1,2}} & \textit{on $I_{2}\setminus I_{1,2}$}\\
\end{array},\right.
\end{equation}
where $I_{1}$, $I_{2}$ are defined in (\ref{I1I2}) and $I_{1,2}$ in (\ref{set_disgunti1}).
\end{remark}
\begin{remark}
For DD-CLS model we considered, DD of $I=\{1,\ldots,n\}\subset \mathbb{N}$ i.e. the index set of columns of m $A$, similarly we can apply DD approach to 2D domain $I\times J\subset \mathbb{N}\times \mathbb{N}$, where $J=\{1,\ldots,(m_0+m_1)\}$ is the rows index set of  $A$. Sub domains obtained are $I_1\times J_1=\{1,\ldots,n_1\}\times\{1,\ldots,m^1\}$ and $I_2\times J_2=\{n_1-s_{I}+1,\ldots,n\}\times\{m^1-s_{J}+1,\ldots,(m_0+m_1)\}$, where $s_{I}, s_{J}\ge 0$ are the number of indexes in common between $I_{1}$ and $I_{2}$, $ J_{1}$ and $J_{2}$,  respectively. Restrictions of $A$ to $I_1\times J_1$ and $I_2\times J_2$ are $A_1:=A|_{I_{1}\times J_1}$ and $A_2:=A|_{I_{2}\times J_2}$.
\end{remark}
\begin{remark}
The cardinality of $J$, i.e. the index set of rows of matrix $A$, represents the number of observations available at time of the analysis, so that  DD of $ I\times J$ allows us to define DD-CLS model after dynamic load balancing of observations by appropriately restricting  matrix $A$.
\end{remark}
\section{DyDD:  Dynamic DD framework}

For effective parallelization of  DD based algorithms,  domain partitioning into sub domains must satisfy certain conditions.  Firstly the computational
load assigned to sub domains must be equally distributed. Usually,  computational cost is proportional to the amount of data  entities assigned to partitions. Good quality partitioning also requires  the volume
of communication during calculation to be kept at its minimum. We employ a dynamic load balancing scheme based on  adaptive and dynamic redefining of initial DD aimed to balance   workload between processors. Redefining of initial partitioning is performed by shifting the boundaries of neighbouring domains (this step is referred to as Migration step). \\
\noindent DyDD algorithm we implement is described by procedure DyDD shown in Table \ref{procedureDyDD}. To the aim of giving a clear and immediate view of DyDD algorithm, in the following  figures (Figures  \ref{fig:three graphs1}-\ref{fig:three graphs3}) we outline algorithm  workout on a reference initial DD configuration made of eight subdomains. We assume that at each point of the  mesh  we have  the value of  numerical simulation result (the so called  background) while the circles denote observations.
DyDD  framework  consists in four steps:

\begin{enumerate}
    \item DD step: starting from the initial partition of $\Omega$ provided by DD-DA framework,  DyDD performs a check of the initial partitioning. If a subdomain is empty, it decomposes  subdomain  adjacent to that domain  which has maximum load (decomposition is performed in $2$ subdomains). See Figure \ref{fig:three graphs1}.
    
    \item Scheduling step: DyDD  computes the amount of observations needed for achieving the average load in each sub domain; this is performed by introducing  a diffusion type algorithm  (by using the connected  graph G associated to the DD) derived by minimizing the Euclidean norm of the cost transfer. Solution of the  laplacian system associated to the graph G gives the amount of data to migrate. See Figure \ref{Fig:LOh14}.
    \item Migration step: DyDD  shifts the boundaries of  adjacent sub domains to achieve a balanced workload. See Figure \ref{fig:three graphs2}.   
    \item Update step: DyDD redefines subdomains such that each one contains the number of observations computed during the scheduling step and it redistributes  subdomains among processors grids. See Figure \ref{fig:three graphs3}.
\end{enumerate}
\noindent Scheduling step is the computational kernel of DyDD algorithm. In particular, it requires definition of laplacian matrix and 
load imbalance associated to initial DD and its solution. Let us give a brief overview of this computation.  Generic element $L_{ij}$ of  laplacian matrix is defined as follows\cite{dynamicload}: 
\begin{equation}
    L_{ij}= \left \{ 
    \begin{array}{ll}
         -1& i \neq j \, and \, edge \, (i,j) \in G  \\
         deg(i)& i=j, \\
         0 & otherwise
    \end{array}
    \right .
\end{equation}

\noindent and the load imbalance $b=\left(l\left(i\right)-\bar{l}\right)$, where $d\left(i\right)$ is the degree of vertex $i$, 
       $l\left(i\right)$ and  $\bar{l}$ are the number of observations and the average workload, respectively.  Hence, as more edges are in G (as the number of subdomains which are adjacent to each other increases)  as more non zero elements are in $L$.         

 Laplacian system $L \lambda = b$, related to the example described below,  is the following:
 
 \begin{equation}
 L= \left [
     \begin{array}{ccccccccc}
          2& -1  & -1 & 0& 0 & 0& 0 &0    \\
         -1& 3 &   -1 & -1 & 0 & 0& 0 &0  \\
         -1& -1 &   4 & -1 & -1 & 0& 0 &0  \\
         0 & -1  &   -1 & 2 & 0 & 0& 0 &0  \\
         0 & 0 &   -1 & 0 & 2 & -1& 0 &0 \\
         0& 0 &   0 & 0  & -1 & 3& -1 &-1 \\
      0& 0 &0  &0 &0 & -1  &2 & -1& \\
             0& 0 &   0 & 0 & 0 & -1& -1 &2 &\\
     \end{array}
     \right ] 
     \end{equation}
     while the right hand side is the vector whose $i$-th component is given by the load imbalance, computed with respect to the average load. In this example, solution of the laplacian system gives
     \begin{equation*}
         \lambda= ( 0.36, 0.25,0.,1.12,-1.,-5.,-6.33,-6.67)
     \end{equation*}
 so that the amount of load (rounded to the nearest integer) which should be  migrated from $\Omega_i$ to $\Omega_j$ is 
 \begin{equation*}
  \begin{array}{c}
 \delta_{1,2}= 1; 
 \delta_{1,3}= 0;
 \delta_{3,2}= 0;
 \delta_{3,4}= 1;
 \delta_{3,5}=1;
  \delta_{5,6}= 2;
  \delta_{6,7}= 0;
  \delta_{6,8}= 1;
  \delta_{7,8}= 1.
 \end{array}
 \end{equation*}
 i.e. $\delta_{i,j}$ is the nearest integer of $ (\lambda_i-\lambda_j)$.
 \begin{figure}
     \centering
     \begin{subfigure}[b]{1\textwidth}
         \centering
         \includegraphics[width=0.45\textwidth]{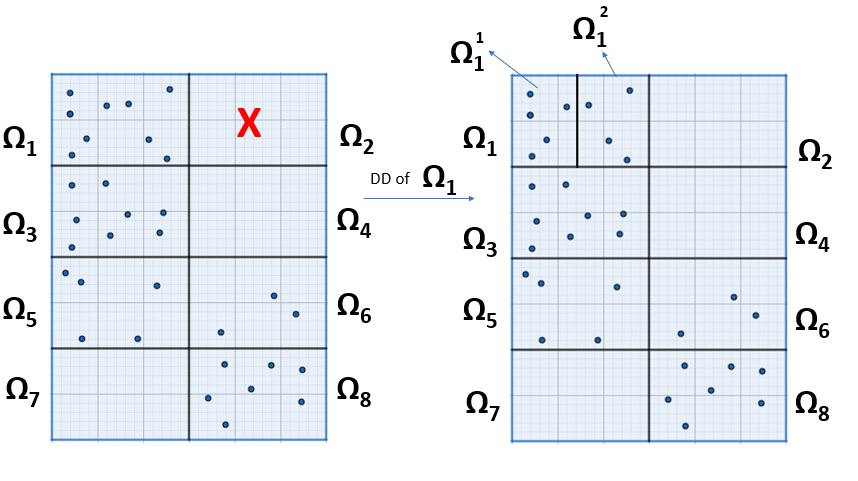}  
         \caption{  $\Omega_1$ is identified as having the maximum load w.r.t. its neighbourhoods.}
         \label{Fig:LOh5}
     \end{subfigure}
     \hfill
     \begin{subfigure}[b]{1\textwidth}
         \centering
         \includegraphics[width=0.45\textwidth]{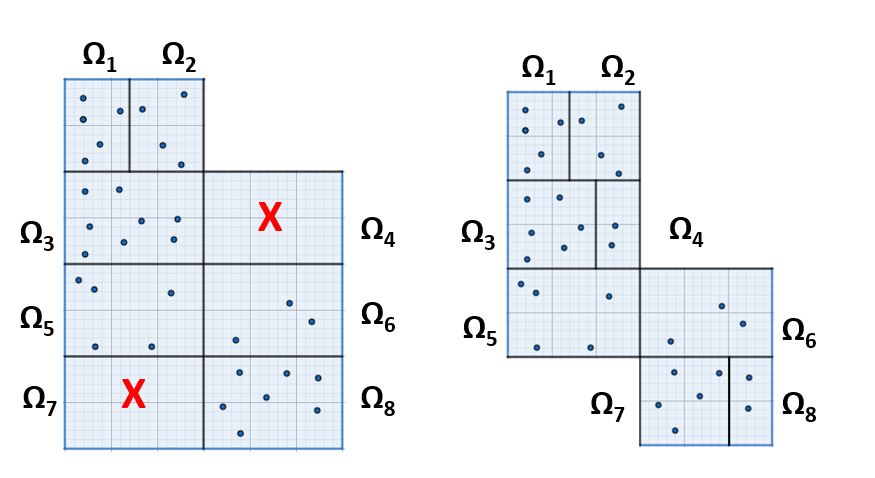} 
         \caption{ $\Omega_4$ and $\Omega_7$ are identified as having the maximum load w.r.t. their neighbourhoods. }
         \label{Fig:LOh6}
     \end{subfigure}
     \hfill
        \caption{DyDD framework - Step 1.  Check of the initial partitioning, identification of subdomains which do not have data or they suffer of any load imbalance and redefinition of  subdomains. We observe that the workload of each subdomain after this re-partitioning   is now $l_{r}(1)= 5$,  $l_{r}(2)= 4$,  $l_{r}(3)= 6$,  $l_{r}(4)= 2$,  $l_{r}(5)= 5$,  $l_{r}(6)= 3$,  $l_{r}(7)= 5$ and  $l_{r}(8)= 2.$ The average load is then $\bar{l}= 4$.  }
        \label{fig:three graphs1}
\end{figure}


\begin{figure}[!ht]
  \centering
  \includegraphics[width=0.5\textwidth]{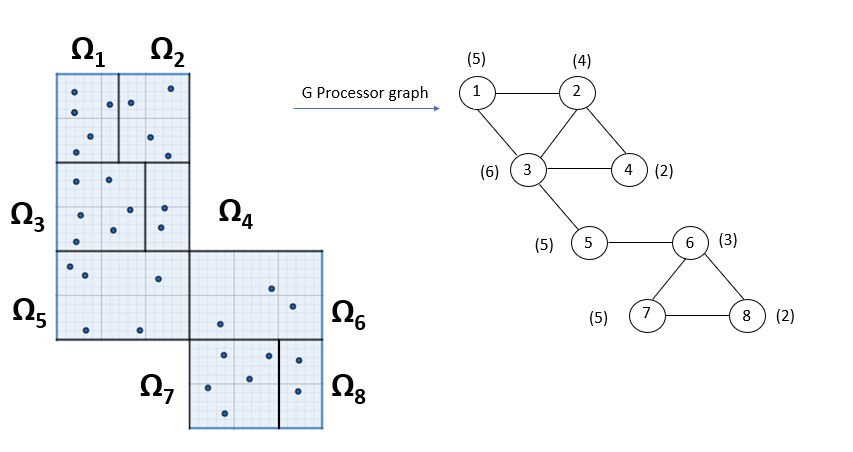}
  \caption{DyDD framework - Step 2. Scheduling. On the right, the graph G associated to the DD of $\Omega$. In brackets the number $l_{r}(i)$ is displayed.}
\label{Fig:LOh14}
\end{figure}

\begin{figure}
     \centering
   \includegraphics[width=0.5\textwidth]{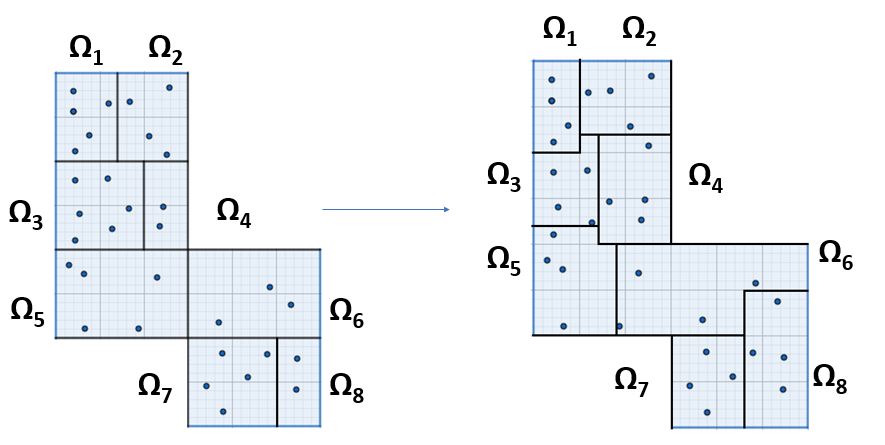}
        \caption{DyDD framework - Step 3. Migration. Redefinition of the boundaries of adjacent subdomains.}
        \label{fig:three graphs2}
\end{figure}

\begin{figure}
\centering
  \includegraphics[width=0.5\textwidth]{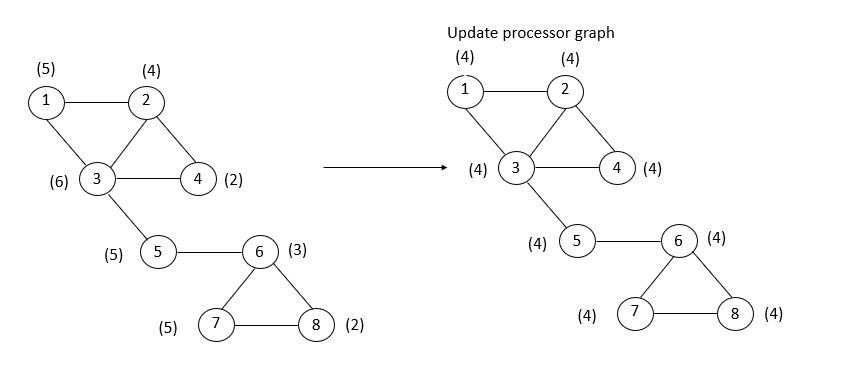} 
        \caption{DyDD framework - Step 4. Update step. Updating of the processor graph. In brackets, the number of observations $l_{fi}(i)$ after DyDD is displayed. We observe that the workload of each subdomain after DyDD is equal to the average load $\bar{l}=4$.}
        \label{fig:three graphs3}
\end{figure}

\section{Validation Results}
Simulations  were aimed to validate the proposed approach by measuring performance  of DyDD algorithm. Performance evaluation was carried out  using
Parallel Computing Toolbox  of  MATLABR2013a on the High Performance hybrid Computing (HPC) architecture of the SCoPE (Sistema Cooperativo Per Elaborazioni scientifiche multidiscipliari) data center, located at  University of Naples Federico II. More precisely, the  HPC architecture is made of $8$ nodes, consisting  of distributed
memory DELL M600 blades connected by a $10$ Gigabit Ethernet technology.
Each blade consists of $2$ Intel Xeon@2.33GHz quadcore processors
sharing $16$ GB RAM memory for a total of 8 cores/blade
and of $64$ cores, in total.  In this case for testing the algorithm we consider up to $n_{sub}=64$ sub domains equally distributed among the cores. This is  an \emph{intra-node}  configuration  implementing a coarse-grained parallelization strategy on multiprocessor systems with many-core CPUs. \\

\noindent DyDD set up. We will refer to the following quantities: $\Omega\subset \mathbb{R}^{2}$: spatial domain; $n=2048$: mesh size; $m:$ number of observations; $p$: number of subdomains and processing units; $i$: identification number of processing unit, which is the same of the associated subdomain; for $i=1,\ldots,p$, 
$deg(i)$: degree of $i$, i.e. number of  subdomains adjacent to $\Omega_i$;
  $i_{ad}(i)\in \mathbb{N}$: identification of  subdomains adjacent to  $\Omega_i$;
   $l_{in}(i)\in \mathbb{N}$: number of observations in $\Omega_i$ before the dynamic load balancing; $l_{r}(i)\in \mathbb{N}$: number of observations in $\Omega_i$ after DD step of DyDD procedure; $l_{fi}(i)\in \mathbb{N}$: number of observations in  $\Omega_i$ after the dynamic load balancing;
$T^p_{DyDD}(m)$: time (in seconds) needed to perform DyDD on $p$ processing units;
 $T_{r}(m)$:  time (in seconds) needed to perform re-partitioning of  $\Omega$;
 $Oh_{DyDD}(m)=\frac{T_r(m)}{T^p_{DyDD}(m)}$  overhead time to the dynamic load balancing. \noindent As measure of  the load balance introduced by  DyDD algorithm, we use:
\begin{equation*}
    \mathcal{E}=\frac{\min_{i}({l_{fi}(i)})}{\max_{i}({l_{fi}(i)})}\\
\end{equation*}

\noindent i.e. we compute  the ratio of the minimum to the maximum of the number of observations of subdomains $\Omega_1,\ldots,\Omega_p$ after DyDD, respectively. As a consequence,  $\mathcal{E}$ = 1 indicates a perfectly balanced system. \\

\noindent Regarding DD-DA, we let
$n_{loc}:=\frac{n}{p}$ be  local problem size and we consider as performance metrics, the following quantities:
 $T^1\left(m,n\right)$
 denoting sequential  time (in seconds) to perform KF solving CLS problem;
 $T_{DD-DA}^{p}\left(m,n_{loc}\right)$
 denoting time (in seconds) needed to perform in parallel DD-KF solving  CLS problem after DyDD; $T_{oh}^p\left(m,n_{loc}\right)$ being the  overhead time (measured in seconds) due to synchronization, memory accesses and communication time among $p$ cores;
$\widehat{x}_{KF}\in \mathbb{R}^n$ denoting KF estimate obtained by applying the KF procedure on CLS problem after DyDD;
$\widehat{x}_{DD-DA}\in \mathbb{R}^n$ denoting DD estimate obtained by applying DD-KF on CLS problem after DyDD;
 $error_{DD-DA}:=\|\widehat{x}_{KF}-\widehat{x}_{DD-DA}\|$ denoting the error introduced by the DD framework; $S^{p}\left(m,n_{loc}\right):=\frac{T^1\left(m,n\right)}{T_{DD-DA}^p(m,n_{loc})}$, which refers to the speed-up of DD-KF parallel algorithm;
$E^{p}\left(m,n_{loc}\right):=\frac{S^p\left(m,n_{loc}\right)}{p}$ which denotes the  efficiency of DD-KF parallel algorithm.
\\

\noindent In the following tables we report results obtained by employing three  scenarios, which are defined such that each one is gradually more articulated than the previous one. It means  that the number of subdomains which are adjacent to each subdomain increases, or the number of observations and  the number of subdomains increase. In this way the workload re distribution increases. \\

\noindent Example 1: First configuration: $p=2$ subdomains and $m=1500$ observations. In {Case1}, both $\Omega_1$ and $\Omega_2$ have data i.e. observations, but they are unbalanced. In  Case2,  $\Omega_1$ has observations and $\Omega_2$ is empty. In Table \ref{tab_Ex1_Case1} and Table \ref{tab_Ex1_Case2}, respectively, we report  values of the parameters after applying DyDD algorithm. This is the simplest configuration we consider just to validate  DyDD framework. In both cases,  $l_{fi}(1)$ and $l_{fi}(2)$, i.e.  number of observations of $\Omega_1$ and $\Omega_2$, are equal to the average load $\bar{l}=750$ and  $\mathcal{E}$ = 1. 
As the workload re distribution of Case 1 and Case 2 is the same, DD-KF performance results of Case 1 and Case 2 are the same, and they are reported in Table \ref{ex1_2},  for $p=2$ only. In Table  \ref{tab_Ex1_execution}  we report performance results of DyDD algorithm.\\
\begin{table}
\centering
\caption{Example 1. DyDD parameters in Case 1. Both subdomains have data but they are unbalanced. We report values of $p$, which is the number of subdomains, $i$ the identification number of processing unit, $deg(i)$  degree of $i$, i.e. number of  subdomains adjacent to $\Omega_i$, $l_{in}(i)$ which is  number of observations in $\Omega_i$ before  dynamic load balancing, $l_{fi}(i)$  number of observations in  $\Omega_i$ after  dynamic load balancing, $i_{ad}$  identification of subdomains adjacent to $\Omega_i$. }
\begin{tabular}{rllllllllll}
\hline
p &  $i$  &  $deg(i)$ & ${l_{in}}$ & $l_{fin}$ & $i_{ad}$  \\ 
\hline
2&1 &1 &{1000} &{750} & 2  \\
 &2 &1 &{500} &{750} & 1  \\
\hline
\end{tabular}
\label{tab_Ex1_Case1}
\end{table}

\begin{table}
\centering
\caption{Example 1. DyDD parameters in Case 2. $\Omega_2$ is empty. We report values of $p$ i.e.  number of subdomains, $i$  identification number of processing unit, $deg(i)$ degree of $i$, i.e.  number of  subdomains adjacent to $\Omega_i$, $l_{in}(i)$ which is  number of observations in $\Omega_i$ before  dynamic load balancing, $l_{r}(i)$  number of observations in $\Omega_i$ after DD step of DyDD procedure, $l_{fi}(i)$  number of observations in  $\Omega_i$ after  dynamic load balancing, $i_{ad}$ which is  identification of subdomains which are adjacent to $\Omega_i$.}
\begin{tabular}{rlllllllllll}
\hline
p   & $i$  &  $deg(i)$ & $l_{in}$ & $l_{r}$&$l_{fin}$  & $i_{ad}$\\  
\hline
2&1 &1 &{1500} &1000&{750} & 2\\
 &2 &1 &{0} &500&{750} & 1\\
\hline 
\end{tabular}
\label{tab_Ex1_Case2}
\end{table}

\begin{table}
\centering
\caption{Example 1. Execution times: we report values of $T^p_{DyDD}(m)$,  time (in seconds) needed to perform DyDD on $p$ processing units,
 $T_{r}(m)$,  time (in seconds) needed to perform re-partitioning of  $\Omega$,
 $Oh_{DyDD}(m)$   overhead time due to  dynamic load balancing and $\mathcal{E}$ measuring load balance.}
\begin{tabular}{rllllllllll}
\hline
\textbf{Case} &  $T^p_{DyDD}(m)$ & $T_r(m)$ & $Oh_{DyDD}(m)$ & $\mathcal{E}$\\ 
\hline
\textbf{1}&$4.11\times 10^{-2}$ &0 & 0 & 1\\
\textbf{2}&$3.49\times 10^{-2}$ &$4.00\times 10^{-6}$ & $1.15\times 10^{-4}$& 1\\
\hline
\end{tabular}
\label{tab_Ex1_execution}
\end{table}

\noindent  Example 2: 
Second configuration. In this experiment we consider  $p=4$ subdomains and $m=1500$ observations, and four cases which are such that the number of subdomains not having observations, increases from $0$ up to $3$. In particular, in {Case 1}, all subdomains have observations. See Table \ref{tab_Ex2_Case1}.  In {Case 2}, only  one subdomain is empty, namely $\Omega_2$. See Table \ref{tab_Ex2_Case2}.  In {Case 3},  two subdomains are empty, namely  $\Omega_1$ and $\Omega_2$ are  empty. See Table \ref{tab_Ex2_Case3}. In  Case 4, three subdomains are empty, namely $\Omega_j$, for $j=1,2,3$, is empty. See Table \ref{tab_Ex2_Case4}. In all cases, $\mathcal{E}$ reaches the ideal value 1 and $l_{fin}(i)=\bar{l}=375$, $i=1,2,3,4$.
Then, DD-KF performance results of all cases are the same and they are reported in Table \ref{ex1_2} for $p=4$. In Table \ref{Ex_2_Executiontimes} we report performance results of the four cases.  \\

\begin{figure}
{\includegraphics[width=.5\textwidth]{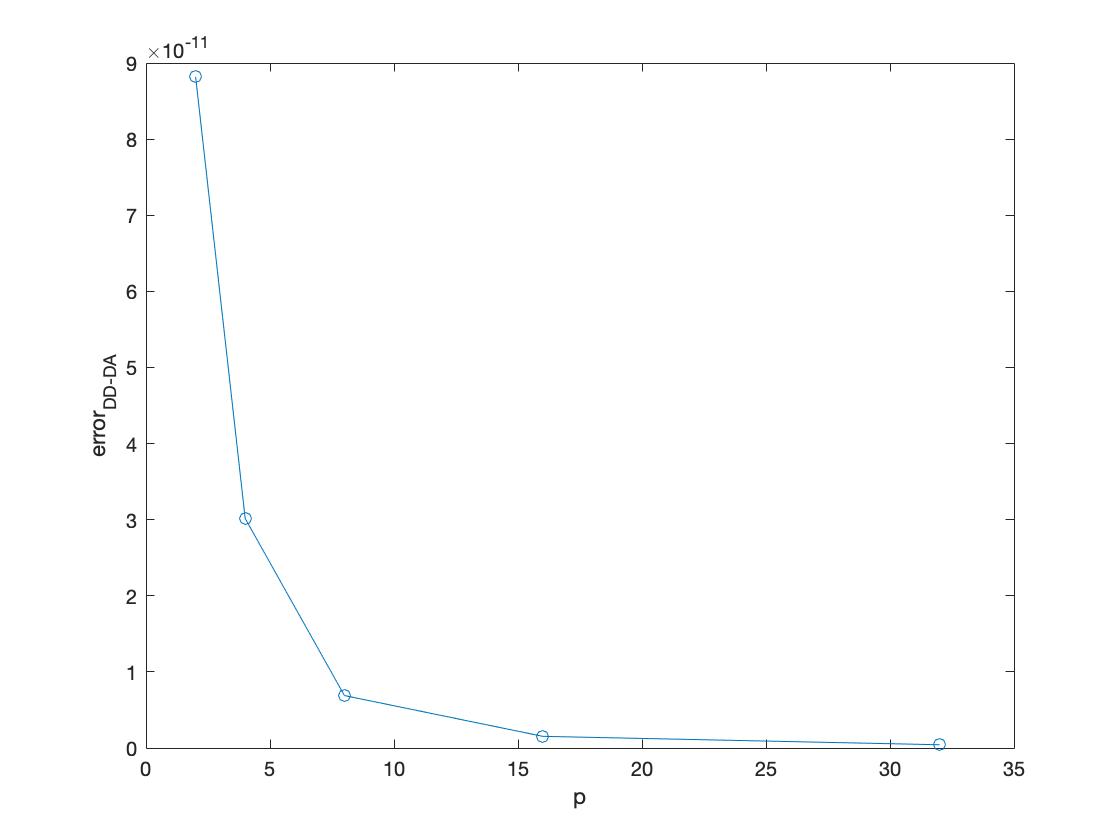}}
\quad
{\includegraphics[width=.5\textwidth]{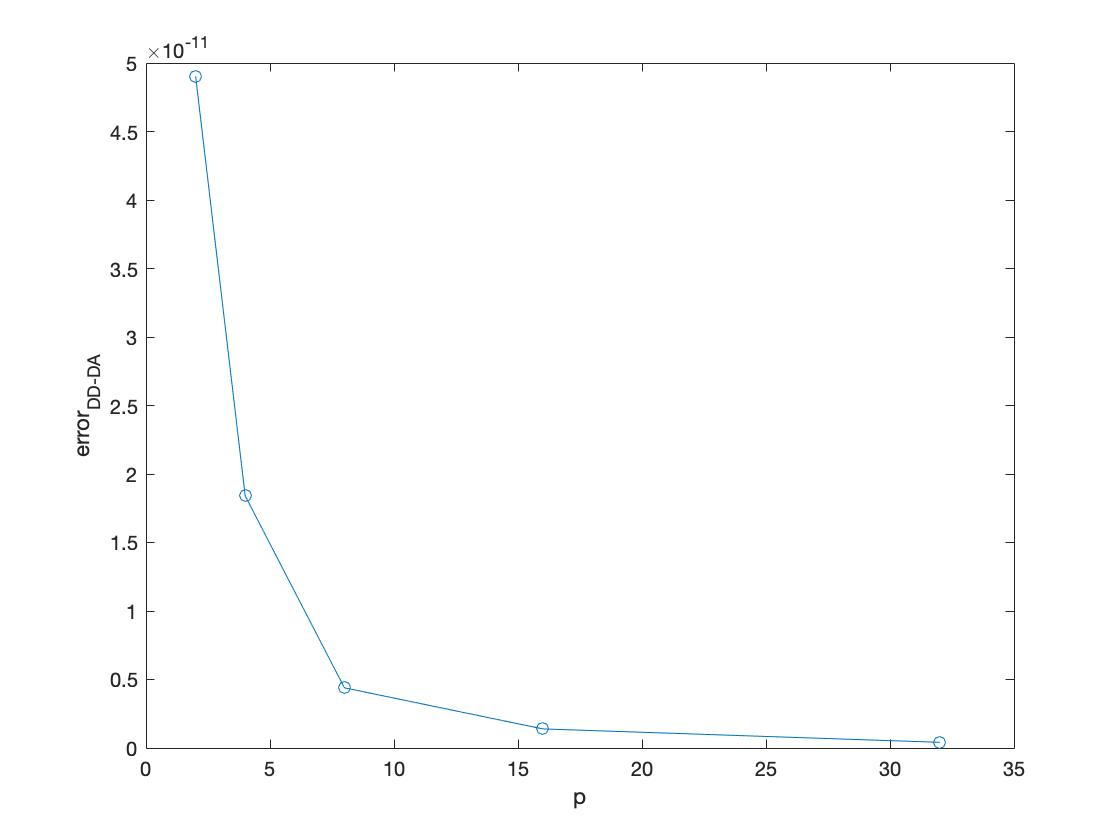}}\\
{\caption{Examples 3 (left)- 4 (right). We report values of $error_{DD-DA}$ versus $p$.}
\label{fig_DD-DA}}
\end{figure}

 \noindent  Example 3. 
We consider $m=1032$ observations and a number of subdomains $p$ equals to  $p=2,4,8,16,32$.  We assume that all subdomains  $\Omega_i$ has observations, i.e. for $i=1,\ldots,p$, $l_{in}(i)\neq 0$;  $\Omega_1$ has $p-1$ adjacent subdomains, i.e. $n_{ad}:=deg(1)=p-1$; $\Omega_i$  has 1 adjacent subdomain i.e. for $i=2,\ldots,p$,  $deg(i)=1$; finally  $i=1,\ldots,p$, we let the  maximum and the  minimum number of observations in $\Omega_i$ be such that  $l_{max}=max_{i}(l_{fin}(i))$ and $l_{min}=min_{i}(l_{fin}(i))$. Table \ref{tab_3} shows performance results  and Figure \ref{fig_DD-DA} reports the error of DD-KF with respect to KF.  \\

\noindent Example 4  
We consider $m=2000$ observations and  $p=2,4,8,16,32$
we assume that $\Omega_i$  has observations, i.e. for  
$i=1,\ldots,p$, $l_{in}(i)\neq 0$; $\Omega_1$ and $\Omega_p$ have 1 adjacent subdomain i.e.  $deg(1)=deg(p)=1$; $\Omega_i$ and $\Omega_p$ have 2 adjacent subdomains i.e. for $i=2,\ldots,p-1$, $deg(i)=2$.  In Table \ref{tab_4} we report performance results and in Figure \ref{fig_DD-DA} the error of DD-KF with respect to KF is shown.  \\

\begin{table}
\centering
\caption{Example 2. DyDD parameters in Case 1. All subdomains have data. We report values of $p$, which is the number of subdomains, $i$  identification number of processing unit, $deg(i)$  degree of $i$, i.e.  number of  subdomains adjacent to $\Omega_i$, $l_{in}(i)$ the number of observations in $\Omega_i$ before  dynamic load balancing, $l_{fi}(i)$ the number of observations in  $\Omega_i$ after  dynamic load balancing, $i_{ad}$  identification of subdomains which are adjacent to $\Omega_i$.}
\begin{tabular}{rlllllllll}
\hline
p &  $i$  &  $deg(i)$ & ${l_{in}}$ & ${l_{fin}}$ & $i_{ad}$  \\ 
\hline
4&1 &2 &{150} &{375}  & [ 2 4 ] \\
 &2 &2 &{300} &{375} & [ 3 1 ] \\
  &3 &2 &{450} &{375}  & [ 4 2 ]  \\
  &4 &2 &{600} &{375}  & [ 3 1 ] \\
\hline 
\end{tabular}
\label{tab_Ex2_Case1}
\end{table}

\begin{table}
\centering
\caption{Example 2. DyDD parameters in Case 2. $\Omega_2$ is empty.  We report values of $p$, which is number of subdomains, $i$ i.e.  identification number of processing unit, $deg(i)$ i.e.  degree of $i$, i.e. number of  subdomains which are adjacent to $\Omega_i$, $l_{in}(i)$ i.e.  number of observations in $\Omega_i$ before  dynamic load balancing, $l_{fi}(i)$ number of observations in  $\Omega_i$ after  dynamic load balancing, $i_{ad}$  identification of subdomains adjacent to $\Omega_i$.}
\begin{tabular}{rlllllllll}
\hline
p   & $i$  &  $deg(i)$ & ${l_{in}}$ & $l_{r}$& ${l_{fin}}$  & $i_{ad}$  \\ 
\hline
4&1 &2 &{450} &450&{375} & [ 2 4 ]  \\
 &2 &2 &{0} &225&{375} & [ 3 1 ]  \\
 &3 &2 &{450} &225&{375} & [ 4 2 ] \\
 &4 &2 &{600} &600&{375} & [ 3 1 ] \\
\hline 
\end{tabular}
\label{tab_Ex2_Case2}
\end{table}

\begin{table}
\centering
\caption{Example 2. DyDD parameters in Case 3. $\Omega_1$ and $\Omega_2$ are empty. We report values of $p$, which is the number of subdomains, $i$ identification number of processing unit, $deg(i)$ i.e. degree of $i$, i.e. number of  subdomains adjacent to $\Omega_i$, $l_{in}(i)$  number of observations in $\Omega_i$ before  dynamic load balancing, $l_{fi}(i)$  number of observations in  $\Omega_i$ after  dynamic load balancing, $i_{ad}$  identification of subdomains which are adjacent to $\Omega_i$.}
\begin{tabular}{rlllllllll}
\hline
p &  $i$  &  $deg(i)$ & ${l_{in}}$ & $l_{r}$&${l_{fin}}$  & $i_{ad}$  \\ 
\hline
4&1 &2 &{0} &{300}&{375}  & [ 2 4 ]  \\
 &2 &2 &{0} &{450}&{375}  & [ 3 1 ]  \\
  &3 &2 &{900} &450&{375}  & [ 4 2 ] \\
  &4 &2 &{300} &600&{375}  & [ 3 1 ]  \\
\hline 
\end{tabular}
\label{tab_Ex2_Case3}
\end{table}

\begin{center}
\begin{table}
\centering
\caption{Example 2. DyDD parameters in Case 4. $\Omega_1$, $\Omega_2$ and $\Omega_3$ are empty. We report values of $p$ i.e.  number of subdomains, $i$   identification number of processing unit, $deg(i)$  degree of $i$, i.e. i.e.  number of  subdomains which are  adjacent to $\Omega_i$, $l_{in}(i)$ the number of observations in $\Omega_i$ before  dynamic load balancing, $l_{fi}(i)$  number of observations in  $\Omega_i$ after  dynamic load balancing and $i_{ad}$   identification of subdomains which are  adjacent to $\Omega_i$.}

\begin{tabular}{rlllllllll}
\hline
p   & $i$  &  $deg(i)$ & ${l_{in}}$ & $l_{r}$& ${l_{fin}}$  & $i_{ad}$   \\ 
\hline
4&1 &2 &{0} &500&{375}  & [ 2 4 ] \\
 &2 &2 &{0} &250&{375}  & [ 3 1 ] \\
 &3 &2 &{0} &250&{375} & [ 4 2 ]  \\
 &4 &2 &{1500} &500&{375}  & [ 3 1 ]  \\
\hline 
\end{tabular}
\label{tab_Ex2_Case4}
\end{table}
\end{center}

\normalsize

\begin{center}
\begin{table}
\centering
\caption{Example2. Execution times: we report values of $T^p_{DyDD}(m)$, i.e.  time (in seconds) needed to perform DyDD algorithm on $p$ processing units,
 $T_{r}(m)$ time (in seconds) needed to perform re-partitioning of  $\Omega$,
 $Oh_{DyDD}(m)$  overhead time to the dynamic load balancing and $\mathcal{E}$ parameter of load balance.}
\begin{tabular}{rlllllllll}
\hline
\textbf{Case}& $T^p_{DyDD}(m)$ & $T_r(m)$ & $Oh_{DyDD}(m)$ & $\mathcal{E}$\\ 
\hline
\textbf{1}& $5.40\times 10^{-2}$ & 0 & 0 &1\\
\textbf{2}& $5.84\times 10^{-2}$ & $2.35\times 10^{-4}$ & $0.4\cdot 10^{-2}$ & 1 \\
\textbf{3}& $4.98\times 10^{-2}$ & $3.92\times 10^{-4}$ & $0.8\cdot 10^{-2}$ & 1\\
\textbf{4}& $4.63\times 10^{-2}$ & $5.78\times 10^{-4}$ &$0.1\cdot 10^{-1}$& 1 \\

\hline 
\end{tabular}
\label{Ex_2_Executiontimes}
\end{table}
\end{center}

\begin{table}[ht]
    \centering
    \caption{Example 1-2: DD-KF performance results in Example 1 and Example 2. We report values of $p$, which is the number of subdomains, $n$ mesh size, $n_{loc}$  i.e. local problem size, $m$ number of observations, $T^1\left(m,n\right)$ sequential  time (in seconds) to perform KF solving CLS problem, $T_{DD-DA}^{p}\left(m,n_{loc}\right)$
 time (in seconds) needed to perform in parallel DD-KF solving  CLS problem with DyDD, $S^{p}\left(m,n_{loc}\right)$ and $E^{p}\left(m,n_{loc}\right)$ the speed-up and efficiency of DD-KF parallel algorithm, respectively.  We applied DyDD to all cases of Example 1 and Example 2 and obtained the perfect load balance, as  reported in Table \ref{tab_Ex1_execution} and Table \ref{Ex_2_Executiontimes}, respectively. As the workload distribution is the same,  DD-KF performance results are the same in all cases of Example 1, then we show results  for $p=2$, only.  In the same way,  for all cases of Example 2, we show results for $p=4$, only. }
    \label{tab:my_label}

\begin{center}
\begin{tabular}{rlllllllllll}
\hline
$p=1$& $n=2048$&$m =1500$ &  &$T^{1}(m,n)=5.67\times 10^{0}$\\ \hline 
$p$ &  $n_{loc}$  &   $T_{DD-DA}^p\left(m,n_{loc}\right)$  & $S^{p}\left(m,n_{loc}\right)$& $E^{p}\left(m,n_{loc}\right)$ \\
\hline
2&1024 &$4.95\times 10^{0}$& $1.15\times 10^{0}$&$5.73\times 10^{-1}$ \\
4&512 &$2.48\times 10^{0}$ &$2.29\times 10^{0}$ & $5.72\times 10^{-1}$  \\
\hline 
\end{tabular}
\label{ex1_2}
\end{center}
\end{table}

\noindent Finally, regarding the accuracy of the DD-DA framework with respect to  computed solution, in Table \ref{ex1_2_b} (Examples 1-2) and in Figure \ref{fig_DD-DA} (Examples 3-4), we get values of $error_{DD-DA}$ . We observe that  the order of magnitude is about  $10^{-11}$ consequently, we may say that  the accuracy of local solutions of DD-DA and hence of local KF estimates, are not impaired by DD approach. \\

\noindent From these experiments, we observe that as the number of adjacent subdomains increases,  data communications required by the workload  re-partitioning among sub domains increases too. Accordingly,  the overhead due   to the load balancing increases (for instance see  Table \ref{Ex_2_Executiontimes}). As expected, the impact of such overhead on the performance of the whole algorithm strongly depends on the problem size and the number of available computing elements. Indeed,  in Case 1 of Example 1 and of Example 2,  when $p$ is small in relation to $n_{loc}$ (see Table  \ref{ex1_2}) this aspect is quite evident. In Example 4, instead, as $p$ increases up to $32$,  and $n_{loc}$ decreases the overhead does not affect  performance results (see Table \ref{tab_4}). In conclusion,  we recognize a sort of trade off between the overhead due to the workload re-partitioning and the subsequent parallel computation. 

\begin{center}
\begin{table}
\centering
\caption{Example 3. Execution times: we report values of $p$, i.e. the number of subdomains, $n_{ad}$, the number of adjacent subdomains to $\Omega_1$, $T^p_{DyDD}(m)$,  time (in seconds) needed to perform DyDD on $p$ processing units, $\mathcal{E}$ which measures  load balance, $l_{max}$ and $l_{min}$ i.e. maximum and minimum number of observations between subdomains after DyDD, respectively. $\mathcal{E}$ depends on  $n_{ad}(\equiv deg(1))$,  i.e. as $n_{ad}(\equiv deg(1))$ increases (consequently $p$ increases), then $\mathcal{E}$ decreases. For $i=1,\ldots,p$, subdomain $\Omega_i$ has observations, i.e. $l_{in}(i)\neq 0$, consequently we do not need to perform re-partitioning of $\Omega$, then  $ T_r(m) \equiv 0$.}
\begin{tabular}{rlllllllll}
\hline
p& $n_{ad}$& $T^p_{DyDD}(m)$ & $l_{max}$ &$l_{min}$& $\mathcal{E}$ \\
\hline
2 & 1 & $6.20\times 10^{-3}$ & $516$ & $515$ & $9.98\times 10^{-1}$\\
4 & 3 & $2.60\times 10^{-2}$ & $258$ & $257$ & $9.96\times 10^{-1}$\\
8 & 7 & $9.29\times 10^{-2}$ & $129$ & $128$ & $9.92\times 10^{-1}$\\
16 & 15 & $1.11\times 10^{-1}$ & $71$ & $64$ & $8.88\times 10^{-1}$\\
32 & 31 & $1.36\times 10^{-1}$ & $39$ & $32$ & $8.21\times 10^{-1}$\\

\hline 
\end{tabular}
\label{tab_3}
\end{table}
\end{center}



\begin{center}
\begin{table}
\centering
\caption{Examples 1-2. We report values of $error_{DD-DA}$, i.e. the error introduced by the DyDD framework, in Example 1 (with $p=2$) and Example 2 (with $p=4$).}
\begin{tabular}{rlllllllll}
\hline
p&$error_{DD-DA}$\\
\hline
2 & $8.16\times 10^{-11}$\\
4 & $8.82\times 10^{-11}$\\
\hline 
\end{tabular}
\label{ex1_2_b}
\end{table}
\end{center}

\begin{table}[ht]
    \centering
    \caption{Example 4. Performance results of DyDD framework: we report values of $p$, whic is the number of subdomains, $n$ i.e. the mesh size, $n_{loc}$  i.e. the local problem size, $m$ the number of observations, $T^1\left(m,n\right)$ i.e. sequential  time (in seconds) needed to perform KF, $T^p_{DyDD}(m)$ i.e. time (in seconds) needed to perform DyDD on $p$ processing units, $T_{DD-DA}^{p}\left(m,n_{loc}\right)$
 i.e. time (in seconds) needed to perform in parallel DD-DA with  DyDD, $S^{p}\left(m,n_{loc}\right)$ and $E^{p}\left(m,n_{loc}\right)$, i.e.  speed-up and efficiency of DD-DA parallel algorithm, respectively.}
    \label{tab:my_label2}
\begin{tabular}{rlllllllllll}
\hline
  $p$& $n=2048$&$m=2000$ &  $T^{1}(m,n)=4.88\times 10^{0}$ & & \\
\hline \hline
 $p$ &  $n_{loc}$  & $T^p_{DyDD}(m)$ & $T_{DD-DA}^p\left(m,n_{loc}\right)$   & $S^{p}\left(m,n_{loc}\right)$& $E^{p}\left(m,n_{loc}\right)$ \\
\hline
2&1024 &$4.10\times 10^{-3}$&$4.71\times 10^{0}$ & $1.04\times 10^{0}$ & $5.18\times 10^{-1}$\\
4& 512&$4.29\times 10^{-2}$&$2.61\times 10^{0}$  &$1.87\times 10^{0}$ &$4.67\times 10^{-1}$\\
8&256 &$1.07\times 10^{-1}$&$8.43\times 10^{-1}$  &$5.79\times 10^{0}$ &$6.72\times 10^{-1}$\\
16& 128&$1.42\times 10^{-1}$& $3.46\times 10^{-1}$ &$1.41\times 10^{1}$& $8.81\times 10^{-1}$\\
32 & 64&$3.49\times 10^{-1}$& $1.66\times 10^{-1}$ &$2.94\times 10^{1}$ & $9.19\times 10^{-1}$\\

\hline 
\end{tabular}
\label{tab_4}
\end{table}

\section{Conclusions}
For effective domain decomposition based parallelization,   the partitioning into sub domains must satisfy certain conditions.  Firstly the computational
load assigned to sub domains must be equally distributed. Usually the computational cost is proportional to the amount of data  entities assigned to partitions. Good quality partitioning also requires the volume
of communication during calculation to be kept at its minimum. In the present work we employed a dynamic load balancing scheme based on  an adaptive and dynamic redefining of initial DD aimed to balance   load between processors according to data location. We call it DyDD.
A  mechanism for dynamically balancing the loads encountered in particular DD
configurations has been included in the  parallel DD framework we implement for solving large scale  DA models. In particular,  we focused on the introduction of a  dynamic redefining of initial  DD in order to deal with  problems  where  the observations are non uniformly distributed and general sparse. This is a quite common issue in Data Assimilation.  
We presented first results obtained by applying DyDD  in space of CLS problems using different scenarios of the initial DD. Performance results confirm the effectiveness of the algorithm. We observed that the impact of  data communications required by the workload  re-partitioning among sub domains affects the performance of the whole algorithm  depending on the problem size and the number of available computing elements. As expected, we recognized a sort of trade off between the overhead due to the workload re-partitioning and the subsequent parallel computation. 
As in the assimilation window the number and the distribution of observations change, the difficulty to overcome is to implement a  load balancing algorithm, which should have to dynamically  allow each subdomain to move independently with time i.e. to balance observations with neighbouring subdomains, at each instant time. We are working on extending DyDD framework to such configurations.


\begin{table}[ht!]
\caption{Procedure DyDD}
\label{procedureDyDD}
\begin{tabular}{l}
\hline
\textbf{Procedure DyDD-Dynamic Load Balancing}(in: $p$, $\Omega$, out: $l_1$,\ldots,$l_p$) \\

\%Procedure DyDD allows to balance observations between adjacent subdomains \\
 \% Domain $\Omega$ is decomposed in $p$  subdomains and  some of them may be empty.\\
\% DBL procedure is composed by: DD step, Scheduling step and Migration Step.\\ 
\% DD step partitions  $\Omega$ in subdomains and if some subdomains have not any observations,
 partitions  adjacent subdomains with maximum load \\ \%in 2 subdomains and redefines the subdomains. \\ 
\% Scheduling step
computes the amount 
 of observations  needed for shifting  boundaries of neighbouring subdomains\\
\%Migration step  decides which
sub domains  should be reconfigured to achieve a balanced load. \\
\% Finally, the Update  step redefines the DD.

\\

\textbf{DD step}\\ 
\% DD step partitions $\Omega$ in 
$ (\Omega_{1},\Omega_{2},\ldots,\Omega_{p})$  \\
\textbf{Define} $n_{i}$, the number of adjacent subdomains of $\Omega_i$  \\

\textbf{Define} $l_i$: the amount of observations in $\Omega_i$ \\
\textbf{repeat}\\
\% identification of  $\Omega_m$, the adjacent subdomain of $\Omega_i$  with the maximum load\\
\textbf{Compute} $l_m=max_{j=1,\ldots, n_{i}}\ (l_{j})$: the maximum amount of observations \\ 

\textbf{Decompose} $\Omega_m$ in 2 subdomains: $\Omega_m\leftarrow (\Omega_m^{1},\Omega_m^{2})$  \\

\textbf{until} ($l_i\neq 0$)\\
\textbf{end of DD Step}\\ \\
\textbf{Begin Scheduling step}\\
\textbf{Define} $G$:  the graph associated with initial partition: vertex $i$ corresponds to  $\Omega_i$\\
\textbf{Distribute} the amount of observations $l_i$ on  $\Omega_i$\\
\textbf{Define} $deg(i)=n_i$, the degree of node $i$ of $G$:  \\
\textbf{repeat}\\
\textbf{Compute} the average load: $\bar{l}=\frac{\sum_{i=1}^{p}l_i}{p}$\\
\textbf{Compute} load imbalance: $b={(l_i-\bar{l})_{i=1,\ldots,p}}$\\
\textbf{Compute} $L$, Laplacian matrix of  $G$\\
\textbf{Call} solve(in:$L,b$, out:$\lambda_i$) \%  algorithm solving the linear system $L \lambda_i= b$\\
\textbf{Compute} $\delta_{i,j}$,  the load increment between the adjacent subdomains $\Omega_i$ and $\Omega_j$.  $\delta_{i,j}$ is the nearest integer of $ (\lambda_i-\lambda_j)$ \\
\textbf{Define} $n_{s_i},n_{r_i}$,  number of those subdomains whose configuration has to be updated  \\ 
\textbf{Update}  graph $G$ \\
\textbf{Update}  amount of observations of $\Omega_i$:  $l_i=l_i-\sum_{j=1}^{n_{s_i}}\delta_{i,j}+\sum_{j=1}^{n_{r_i}}\delta_{j,i}$\\
\textbf{until} $(max \|l_i-\bar{l}\|==\frac{deg(i)}{2})$ \% i.e. maximum load-difference is $deg(i)/2$\\
\textbf{end Scheduling step}\\ \\
\textbf{Begin Migration Step}\\
\textbf{Shift}   boundaries of two adjacent sub domains  in order to achieve a balanced load.\\
\textbf{end Migration Step}\\ \\
\textbf{Update} DD of $\Omega$\\ \\
\textbf{end Procedure  DyDD}\\
\hline
\end{tabular}
\end{table}

 \begin{wrapfigure}{l}{30mm} 
    \includegraphics[width=1in,height=1.7in]{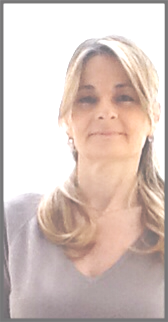}
  \end{wrapfigure}\par
  \textbf{Luisa D'Amore} Luisa D'Amore has  the degree in Mathematics, and the Ph.D. in Applied Mathematics and Computer Science. She is  professor of Numerical Analysis at University of Naples Federico II. She is member of the Academic Board of the Ph.D. in Mathematics and Applications, at University of
Naples Federico II where she teaches courses of Numerical Analysis, Scientific Computing and Parallel Computing. Research activity
is placed in the context of Scientific Computing. Her main interest is devoted to  designing  effective numerical algorithms solving ill-posed inverse problems arising in the applications, such as image analysis, medical imaging, astronomy, digital restoration of films and data assimilation. The need of computing the numerical solution within a suitable time,  requires the use of advanced computing architectures. This involves designing and development of parallel algorithms and software capable of exploiting the high performance of emerging computing infrastructures.
Research produces a total of about 200 publications in refereed journals and conference proceedings.\\

\begin{wrapfigure}{l}{30mm} 
    \includegraphics[width=1in,height=1.5in]{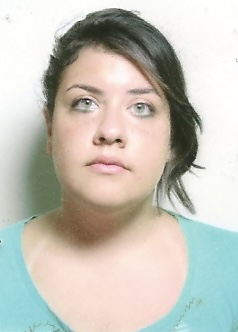}
  \end{wrapfigure}\par
  \textbf{Rosalba Cacciapuoti} Rosalba Cacciapuoti received the degree in Mathematics at University of Naples Federico II. She is a student of the  PhD course in Mathematics and Applications at the University of Naples, Federico II. Her research activity is focused on designing of  parallel algorithms for solving Data Assimilation problems.  

\clearpage
\nocite{*}

\begin{thebibliography}{9}


\bibitem{DDTV}L. Antonelli, L. Carracciuolo,  M. Ceccarelli, L. D'Amore, A. Murli - Total Variation Regularization for Edge Preserving 3D SPECT Imaging in High Performance Computing Environments,  Sloot P.M.A., Hoekstra A.G., Tan C.J.K., Dongarra J.J. (eds) Computational Science,  ICCS 2002. ICCS 2002. Lecture Notes in Computer Science, vol 2330. Springer, Berlin, Heidelberg


\bibitem{JCP} R. Arcucci, L. D'Amore, J. Pistoia, R. Toumi, A.Murli, On the variational data assimilation problem solving and sensitivity analysis, Journal of Computational Physics, 335, pp.311-326, 2017. 

\bibitem{JPP} Arcucci, R., D'Amore, L., Carracciuolo, L., Scotti, G., Laccetti, G. (2017). A Decomposition of the Tikhonov Regularization Functional oriented to exploit hybrid multilevel parallelism. INTERNATIONAL JOURNAL OF PARALLEL PROGRAMMING, vol. 45, p. 1214-1235, ISSN: 0885-7458, doi: 10.1007/s10766-016-0460-3

\bibitem{WCEAS} Arcucci, R.,  D'Amore, L.,  Carracciuolo, L., On the problem-decomposition of scalable 4D-Var Data Assimilation model, Proceedings of the 2015 International Conference on High Performance Computing and Simulation, HPCS 2015
2 September 2015,  Pages 589-594, 13th International Conference on High Performance Computing and Simulation, HPCS 2015; Amsterdam; Netherlands; 20 July 2015 through 24 July 2015.


\bibitem{PPAM2017}  Arcucci, R., D'Amore, L., Celestino, S.,  Laccetti, G., Murli, A. (2016). A Scalable Numerical Algorithm for Solving Tikhonov Regularization Problems. In: Parallel Processing and Applied Mathematics. LECTURE NOTES IN COMPUTER SCIENCE, vol. 9574, p. 45-54, HEIDELBERG:SPRINGER, ISBN: 978-3-319-32152-3, ISSN: 0302-9743, doi: 10.1007/978-3-319-32152-3-5

\bibitem{Battistelli} G. Battistelli, L. Chisci,  Stability of consensus extended Kalman filter for distributed state, Automatica
Volume 68, June 2016, pp. 169-178



\bibitem{Boillat} J. E. Boillat - Load balancing and Poisson equation in a graph, Concurrency: Practice and Experience, 2, 289-313, 1990.


\bibitem{deeplearning4} M. M. Bronstein, J. Bruna, Y. LeCun, A. Szlam, P. Vandergheynst, Geometric deep learning:
going beyond Euclidean, IEEE SIG PROC MAG, arXiv:1611.08097v2 [cs.CV],  3 May 2017


\bibitem{Cacuci} D.G. Cacuci. Sensitivity and Uncertanty Analysis , Chapman, 
Hall/Crc, NY, 2003

\bibitem{Chan} T. F. Chan, T. P. Mathew, Domain Decomposition algorithms, Acta Numerica, 1994, pp. 61-143.


\bibitem{Cybenco} G. Cybenco - Dynamic load balancing for distributed memory multiprocessors, Journal of Parallel and Distributed Computing, 7, 279-301, 1989.

\bibitem{Cohn} S. E. Cohn, An introduction to estimation theory, J. Meteor. Soc. Japan, 75 (1B) (1997), pp. 257-288.


\bibitem{GaussianF} S. Cuomo, G. Severino,  A. Sommella, G. D'Urso, Numerical Effects of the Gaussian Recursive Filters in Solving Linear Systems in the 3Dvar Case Study, Water Resources Research, 53 (10), pp. 8614-8625. DOI: 10.1002/2017WR020904, 2017.





\bibitem{4DVAR} L. D'Amore, R. Cacciapuoti, Model Reduction in Space and Time for decomposing ab initio 4D Variational Data Assimilation Problems, Applied Numerical Mathematics, 2021, Volume  160, pp. 242-264 Elsevier, https://doi.org/10.1016/j.apnum.2020.10.003.
\bibitem{DD-KF} D'Amore, L., Cacciapuoti, R., V. Mele - Ab initio Domain Decomposition Approaches for Large Scale Kalman Filter Methods:
a case study to Constrained Least Square Problems, 13th International Conference, PPAM 2019, Bialystok, Poland,
September 8-11, 2019, 10.1007/978-3-030-43222-5, LNCS Vol. 12044, Springer.


\bibitem{Diniz} P. Diniz, S. Plimpton, B. Hendrickson, and R. Leland, Parallel algorithms for dynamically partitioning unstructured grids, in Proc. 7th SIAM Conf. Parallel Processing for Scientific Computing, SIAM, 1995, pp. 615-620.


\bibitem{Horton} G. Horton - A multi-level diffusion method for dynamic load balancing. Parallel Computing, 9, 209-218, 1993.

\bibitem{dynamicload} Y.F. Hu, R.J. Blake and D.R. Emerson - An optimal migration algorithm for dynamic load balancing,
Concurrency: Practice and Experience 10(6):467-483, 1998.

\bibitem{efficienza} G.A. Kohring - Dynamic load balancing for parallelized particle
simulations on MIMD computers, Parallel Computing 21 pp.683-693, Elsevier, 1998.



\bibitem{Ricerche2019} D'Amore, L.,  Cacciapuoti, R., A note on domain decomposition approaches for solving 3D variational data assimilation models, 
Ricerche di Matematica, 2019.doi:10.1007/s11587-019-00432-4


\bibitem{JSC} D'Amore, L., Arcucci, R., Carracciuolo, L.,   Murli, A. (2014). A scalable approach for Variational Data Assimilation. JOURNAL OF SCIENTIFIC COMPUTING, vol. 61, p. 239-257, ISSN: 0885-7474, doi: 10.1007/s10915-014-9824-2



\bibitem{JCOM2015}  L. D'Amore, G. Laccetti, D. Romano, G. Scotti, A. Murli - Towards a parallel component in a GPU–CUDA environment: a case study with the L-BFGS Harwell routine, International Journal of Computer Mathematics Volume 92, 2015. Issue 1, https://doi.org/10.1080/00207160.2014.899589

\bibitem{CAI2019} L. D'Amore, V. Mele, D. Romano, G. Laccetti, D. Romano, A Multilevel  Algebraic Approach for Performance Analysis of Parallel Algorithms, Computing and Informatics, 38 (4), DOI:10.31577/cai$\_$2019$\_$4$\_$817, 2019


\bibitem{Evensen} G. Evensen, The ensemble Kalman filter: Theoretical formulation and practical implementation, Ocean Dynam. 53 (2003) 343-367.


\bibitem{FUJIMOTO} M. Fujimoto, M. Kawahara, Domain Decomposition for Kalman Filter
Method and Its Application to Tidal Flow at
Onjuku Coast, Proceedings of 12th International Conference on Domain Decomposition Methods, 2001
Editors: Tony Chan, Takashi Kako, Hideo Kawarada, Olivier Pironneau, ISBN 4-901404-00-8.


\bibitem{Gander} M. J. Gander, Schwarz methods over the course of time, ETNA, 31:228-255, 2008.

\bibitem{least1} W. Gander, Least squares with a quadratic constraint,
Numerische Mathematik, vol. 36, pp. 291-307, 1980.


\bibitem{Homescu} C. Homescu, L. R. Petzold, R. Seban,  Error Estimation for REduced Order Models of Dynamical Systems, UCRL-TR-2011494, December 2003.

\bibitem{Kalman60} R. E. Kalman, A New Approach to Linear Filtering and Prediction Problems, Transaction of the ASME - Journal of Basic Engineering, pp. 35-45, 1960.

\bibitem{Kalnay} E. Kalnay,  Atmospheric Modeling, Data Assimilation and Predictability Cambridge University Press, 2003

\bibitem{Kahn} U. A. Khan, Distributing the Kalman Filter for
Large-Scale Systems, IEEE TRANSACTIONS ON SIGNAL PROCESSING, VOL. 56, NO. 10, OCTOBER 2008




\bibitem{Concurr} A. Murli, L. D'Amore, G. Laccetti, F. Gregoretti, G. Oliva,  A multi-grained distributed implementation of the parallel Block Conjugate Gradient algorithm, Concurrency Computation Practice and Experience
Volume 22, Issue 15, October 2010, Pages 2053-2072


\bibitem{Fourier2002} A. Murli, L. D'Amore. Regularization of a Fourier series method for the Laplace transform inversion with real data, Inverse Problems, Vol 18(4), 2002.

\bibitem{Nichols} N. K. Nichols,  \emph{Mathematical concepts of data assimilation}. In: Lahoz, W., Khattatov, B. and Menard, R. (eds.) Data assimilation: making sense of observations. Springer, pp. 13-40, 2010.


\bibitem{Nocedal} J. Nocedal, S.J. Wright - Numerical Optimization, Springer-Verlag, 1999.



\bibitem{Rozier} D. Rozier, F. Birol, E. Cosme, P. Brasseur,J. M. Brankart, J. Verron, A Reduced-Order Kalman Filter for Data Assimilation in Physical Oceanography, SIAM REVIEW, Vol. 49, No. 3, pp. 449-465, 2007


\bibitem{sorenson} 
H. W. Sorenson, Least square estimation: from Gauss to Kalman, IEEE Spectrum, Vol. 7, pp. 63-68, 1970.


\bibitem{Schwarz} H.A. Schwarz.  Journal fur die reine und angewandte Mathematik, 70:105-120, 1869.



\bibitem{Xu1} C. Z. Xu and F. C. M. Lau - Analysis of the generalizes dimension exchange method for dynamic load balancing, Journal of Parallel and Distributed Computing, 16, 385-393, 1992.

\bibitem{Xu2} C. Z. Xu and F. C. M. Lau - The generalized dimension exchange method for load balancing in K-ary ncubes and variants, Journal of Parallel and Distributed Computing, 24, 72-85, 1995. 


\end{thebibliography}
\end{document}